\documentclass[conference]{IEEEtran}
\IEEEoverridecommandlockouts

\usepackage{cite}
\usepackage{amsmath,amssymb,amsfonts}
\usepackage{algorithmic}
\usepackage{graphicx}
\usepackage{textcomp}  
\usepackage{xcolor}
\def\BibTeX{{\rm B\kern-.05em{\sc i\kern-.025em b}\kern-.08em
    T\kern-.1667em\lower.7ex\hbox{E}\kern-.125emX}}
\begin{document}

\title{An ontological analysis of risk in Basic Formal Ontology\\}

\author{\IEEEauthorblockN{Federico Donato}
\IEEEauthorblockA{\textit{Department of Philosophy SUNY at Buffalo} \\
\textit{NCOR}\\
Buffalo, USA \\
fdonato@buffalo.edu}
\and

\IEEEauthorblockN{Adrien Barton}
\IEEEauthorblockA{\textit{IRIT,  CNRS} \\
\textit{Université de Toulouse}\\
Toulouse, France\\
adrien.barton@irit.fr}
}

\maketitle

\begin{abstract}
The paper explores the nature of risk, providing a characterization using the categories of the Basic Formal Ontology (BFO). It argues that the category Risk is a subclass of  BFO:\textit{Role}, contrasting it with a similar view classifying Risk as a subclass of BFO:Disposition. This modeling choice is applied on one example of risk, which represents objects, processes (both physical and mental) and their interrelations, then generalizing from the instances in the example to obtain an overall analysis of risk, making explicit what are the sufficient conditions for being a risk. Plausible necessary conditions are also mentioned for future work.

\end{abstract}

\begin{IEEEkeywords}
ontology, risk, BFO, role, disposition
\end{IEEEkeywords}

\section{Risk Databases and Ontologies}
 
\subsection{The need for a risk ontology}

Risks are ubiquitous. Your likelihood of receiving a loan and the interest rates associated with the loan are directly correlated to the risk the bank sees in losing its money if you fail to repay it. Companies in the stock market are evaluated by brokers and investors based on their risk profile. Policymakers ponder whether a hydrogeological plant will put the environment at risk and how this will negatively affect their net political campaign. Physicians recommend cost-benefit analyses to their patients for surgical operations that could potentially extend their lives, but also pose risks. Risks can be severe or minor and present in the population according to various statistical distributions. Risky situations, such as in a terrorist emergency, can arguably exist even if the terrorist attack is just suspected but never occurs.

These plausible natural language descriptions of events related to risk can be found in terminologies, contracts, and many other documents. Moreover, in today's information-driven society, risks are stored and analyzed using databases and risk analysis equations. The databases focusing on this specific domain are called risk registers, and they can track all kinds of risks, connecting them to people, processes, economical assets, medical status, locations and times \cite{b1}. However, the conceptual foundations of risk lack clarity and there are ongoing discussions in both the natural language and database communities regarding risk \cite{b2}. Despite the prevalence of risk discussions, there is no universally accepted definition among all scholars or institutions, which hinders effective action to prevent or mitigate risk. The confusion is exacerbated by the numerous complex vocabulary terms and database fields that seem to be linked to or depend on risk, such as “risk propagation”, “risk prevention”, “risk aggregation”, “vulnerability” and “cybersecurity”, among many others \cite{b3} \cite{b4}. In fact, precise knowledge of what is at risk, what causes risk, and how many risks there are is essential for an intervention that is timely and proportional to the prospected damage.

Questions that still pertain to the contemporary theory of risk include: What is the nature of risks (material entities, beliefs, special kinds of probabilities)? Is the risk of losing the reputation of my company identical to the opportunity of a rival company to gain reputation? Does a risk that never realizes actually exist? Is risk objective or subjective, absolute or relative? As a preliminary to address these questions, we will propose an explicit and logically-based definition or risk based on top-level ontology categories.

\section{Definitions of risk and their elements }

Ontologies are information artifacts that specify meanings of terms and capture our knowledge of the world, categorize the relevant entities of a domain in hierarchies, leverage logical deduction, and enable interoperability between diverse databases \cite{b5}. We align our framework with the Basic Formal Ontology \cite{b6}, a widely used top-level ontology that provides common ground for building domain ontologies and establishing a common semantics across diverse databases \cite{b7}. To our knowledge, only two other risk ontologies have been aligned with top-level ones, indicating that the research field is still relatively underdeveloped \cite{b8}  \cite{b9}. This paper will argue step by step for the following sufficient condition on being a risk: \\

\noindent \textit{If}: 
(i) r is a role \& \\
(ii) r is externally grounded in an aversion that an agent or group of agents have towards the possible realizations of the role, \\
\textit{then} r is a risk. \\

In order to understand risk, we need to define the terms that are involved in   the condition, namely something that is risky or the source of risk, something that is put at stake or at risk, something that is valuable or desirable, and some process that might involve uncertainty, as risk is commonly understood \cite{b10}. The implication of values, interests or desires in the definition of risk might raise questions on the subjective or objective nature of risks \cite{b11} \cite{b12}. We will consider the interplay between subjectivity and objectivity, especially with the relation of external grounding. 

To develop a comprehensive understanding of risk, we aim to create an Aristotelian definition, that categorizes the entity within a genus, namely a superclass, and using a differentia, which explains how it differs from other entities within the same genus, and clarifies its ontological relations with other entities \cite{b13}. Structuring a definition in an Aristotelian way helps translating it in a logical form, enabling automatic reasoning. 

This short paper proposes an exploration in a new direction that the ontology literature has not considered so far. The paper is structured as follows: the third section contrasts two conceptions of risks as BFO:\textit{Disposition} and BFO:\textit{Role}; the fourth section considers the subjective side of risk, involving a mental attitude of an agent and the target of this mental attitude; the conclusion provides a way of categorizing risk with sufficient conditions, also listing further works to expand the ontology and limitations of the present ontological model.

\section{Risks as Realizable Entities}   

Having introduced a general picture of what risk plausibly is and some open questions about it, we now turn to a unified definition of risk. In this section, we propose a view of risk as a realizable and externally grounded property. In this formulation, externality is understood as involving two or more mereologically distinct entities \cite{b14}, while grounding is a relation between entities in which one exists because of the other \cite[p.~100]{b6} \cite{b15}. We will examine two views that meet those requirements: risk as a disposition and risk as a role, ultimately favoring the latter. Both views on risk are developed in the BFO framework, which is briefly explained in the following.

BFO is a top-level ontology \cite{b6} that identifies two very general classes of entities, continuants and occurrents. Continuants can endure in time, existing at different times and changing through them, whereas occurrents are temporally extended and have temporal parts, but cannot change in time. Examples of continuants include an orange and its mass of 200g, both of which exist for a certain period of time. In contrast, occurrents include the ripening process of the orange and the time span during which it occurs. Continuants encompass various entities, including material entities and properties, named “specifically dependent continuants” (SDC) in BFO. SDCs include qualities, such as the orange’s mass, as well as latent properties that can be realized in a process. Those latent properties, called “realizable entities”, include dispositions, that depend solely on the physical make-up of the entity. For instance, an orange has the disposition to dry out under the hot sun, even when it is fully hydrated and no drying process is occurring. This disposition exists independently of the orange's current state. They also include properties that do not depend only on the physical make-up of the entity, like the role of this orange of being the winner in a fruit competition, after being chosen by a jury. More specifically, BFO defines dispositions and roles as follows \cite{b15}:\\

d is a disposition $=_{def}$ (i) d is a realizable entity \& (ii) d’s bearer is some material entity \& (iii) d is such that, if it ceases to exist, then its bearer is physically changed \& (iv) d’s realization occurs because this bearer is in some special physical circumstances \& (v) this realization occurs in virtue of the bearer’s physical make-up. \\

r is a role $=_{def}$ (i) r is a realizable entity \& (ii) r exists because there is some single bearer that is in some special physical, social, or institutional set of circumstances in which this bearer does not have to be \& (iii) r is not such that, if it ceases to exist, then the physical make-up of the bearer is thereby changed.     \\

To illustrate the analysis of risks within the BFO framework, consider the following scenario: Paul is walking under a ceiling (named ceiling$_{0}$) that is unstable and might fall over him. Paul is a BFO:\textit{Object} and ceiling$_{0}$ a BFO:\textit{Material entity}, which bears a disposition instability$_{0}$ that might be realized in a process of collapsing (in which case it will be called collapsing$_{0}$), which has as participant ceiling$_{0}$. If it is undesirable for Paul that ceiling$_{0}$ would collapse on him, then a corresponding risk arises when he is in the room. This scenario raises a fundamental question: to which ontological category do individual risks belong? We will examine two competing views: the first \cite{b17} posits that risks are dispositions, while this paper will propose an alternative view according to which they are roles. We will present both views in order to compare their strengths and weaknesses.

\subsection{First Alternative: Risks as Dispositions}

Grenier and Barton \cite{b17} argue that risks are dispositions whose realizations are undesirable for certain agents. For every agent x$_{i}$, they introduce the class of dispositions whose possible realizations are undesirable for x$_{i}$,  named here \textit{Risk$^{d}$-for-x$_{i}$}. Then, they introduce a class (written here \textit{\textit{Risk$^{d}$}}) which is the union class of the classes \textit{Risk$^{d}$-for-x$_{i}$} for all i (thus, \textit{Risk$^{d}$-for-x$_{i}$} is a subclass of \textit{Risk$^{d}$} which is a subclass of BFO:\textit{Disposition}; note that since  \textit{Risk$^{d}$-for-x$_{i}$} explicitly mentions an individual x$_{i}$, it is arguably not a \textit{bona fide} universal but a defined class or collection of particulars, as introduced by Smith and Ceusters \cite{b18}).

According to this view, a disposition d$_{0}$ can be an instance of \textit{Risk$^{d}$-for-x$_{i}$} at one time but not at another, depending on the agent x$_{i}$’s attitude towards the possible realizations of d$_{0}$ at those times: more precisely, depending on whether x$_{i}$ finds d$_{0}$’s possible realization undesirable or not. In the above mentioned example, when Paul leaves the room and becomes indifferent to the possible collapse of ceiling$_{0}$, instability$_{0}$ is no longer an instance of \textit{Risk$^{d}$-for-Paul}, although it continues to exist and is still a disposition. This conception attempts to capture the dual nature of risk: the dispositional nature of an instance of \textit{Risk$^{d}$} captures the objective side of risk, whereas the fact that this disposition instantiates \textit{Risk$^{d}$-for-Paul} is based on Paul’s desires and reflects the subjective nature of risk, thus resolving one of the questions we posed in the introduction.

Let’s revisit two features of risks that must be taken into account: realizability and external grounding. In this view, risk is a disposition, and thus a realizable entity. A disposition is a risk (or not) based on external entities, namely the desires of some agent; thus, although its existence does not depend on such external entities (instability$_{0}$ exists independently of the mental attitudes of Paul), its risk status (the fact that it instantiates \textit{Risk$^{d}$}) does. To summarize, in this view, a disposition that is a risk is internally grounded in the sense that it exists independently of the external world (see \cite{b19} for more considerations on grounding); but its risk status (that is, the fact that it instantiates the class \textit{Risk$^{d}$}) is externally grounded, as it is contingent upon external factors. To illustrate this distinction, consider a similar example. The rectangular shape of this table (a quality) is not externally grounded, but it could instantiate the class \textit{Shape-seen-by-Paul} (which is, again, a defined class and not a \textit{bona fide} universal) based on external features (namely Paul seeing this shape). Therefore, in this view, a disposition is not a risk essentially but accidentally: a disposition can become a risk, and a risk can lose its risk status and become a mere disposition, based on the desires of agents. Note also that in the dispositional approach, collapsing$_{0}$ is an instance of a more specific subclass than BFO:\textit{Process}, in this case an instance of the defined class \textit{Process-undesirable-for-Paul}. This instantiation is accidental, similarly to the accidental character of the instantiation of \textit{Risk$^{d}$-for-Paul} by instability$_{0}$. Fig. 1 summarizes this approach:

\begin{figure}[htbp]
\begin{center}
\includegraphics[scale=0.160]{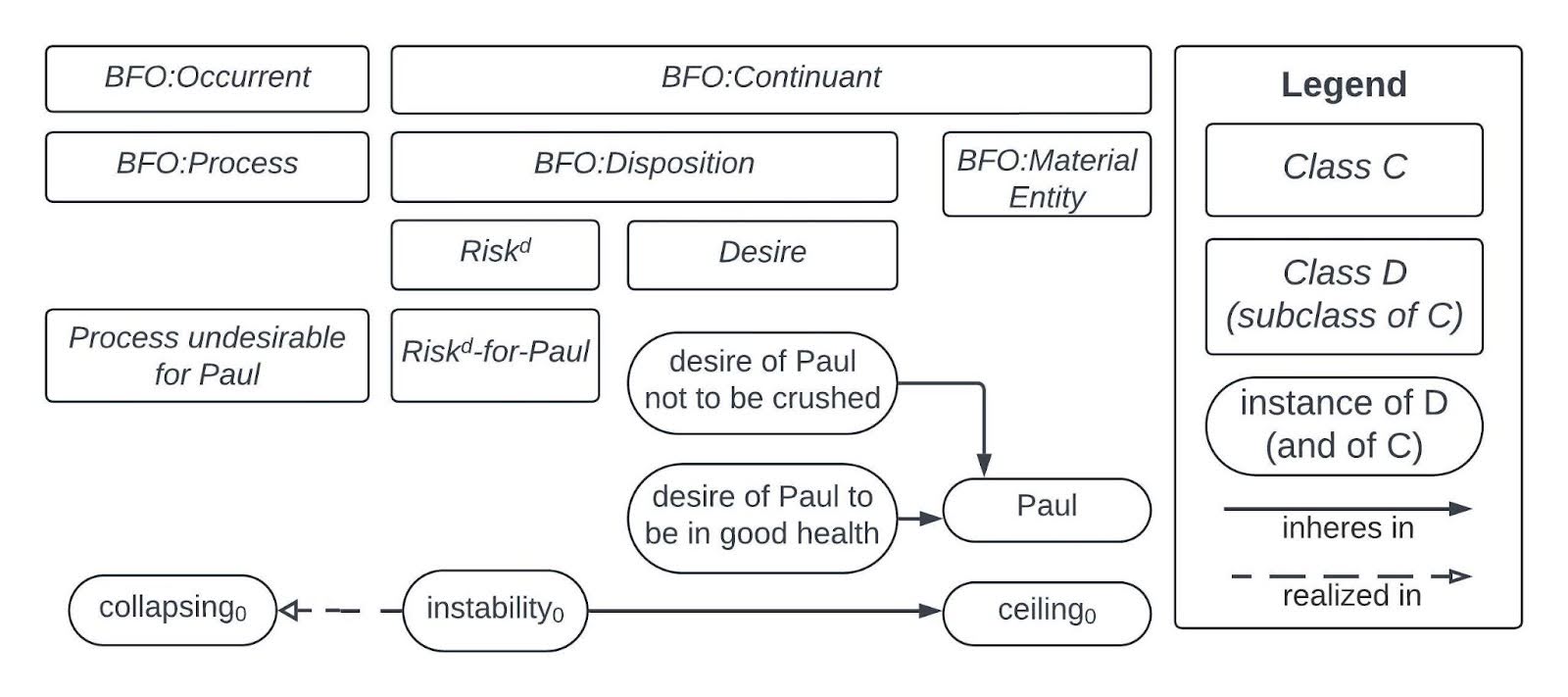}
\end{center}
\caption{\textit{Risk$^{d}$} as a BFO:\textit{Disposition}}
\label{fig}
\end{figure}

\subsection{Second Alternative: Risks as roles}

Let us now contrast this dispositional approach with a new approach according to which risks are another kind of realizable entity, namely roles. To contrast the two views, a risk according to the dispositional approach will be called a “dispositional risk”, and a risk according to the role approach will be called a “risk role”. 

To put it informally, roles are SDCs that existentially depend on the context in which they are present. Instead of having a disposition that starts and stops instantiating \textit{Risk$^{d}$-for-Paul} at different times, we can think of an instance of a class named \textit{Risk$^{r}$}, itself a subclass of BFO:\textit{Role}, which remains a risk throughout its existence. As in the dispositional approach, the role approach considers a risk as a BFO:\textit{Realizable Entity}, since it can be activated and realized. It also agrees with the dispositional approach in considering the source of risk as ceiling$_{0}$, a material entity bearing properties relevant for the risky situation. In the ceiling example, we will say that ceiling$_{0}$ bears a risk role hazard$_{0}$ when Paul is located under it and wants to preserve his health.

As in the first alternative, what makes ceiling$_{0}$ risky for Paul is external to ceiling$_{0}$. Additionally, ceiling$_{0}$ does not bear the risk role hazard$_{0}$ by necessity: it does so because of the context, which consists of Paul being in the vicinity of the ceiling, desiring not to be hurt, and of ceiling$_{0}$ being unstable. Expanding the scenario, if Vladimir, an enemy of Paul, wanted him crushed, then the instability of ceiling$_{0}$ would not be a risk for Vladimir, but an opportunity for him. So, ceiling$_{0}$ does not bear a risk on its own, independently of the rest of the world, but due to the mental attitudes of some agents. Similarly, in the dispositional approach, instability$_{0}$ is an instance of \textit{Risk$^{d}$} based on external features, namely in virtue of the desires of Paul.

The dispositional approach and the role approach have similarities and differences in how they deal with this externally-grounded  character of risk. In both approaches, as we saw, a material entity bears a dispositional risk or a risk role on the basis of external features. According to the dispositional approach, as explained earlier, a disposition can become or stop being a risk - similarly to a human who can continue to exist as he grows up even if he is not a child anymore. For example, instability$_{0}$ continues to exist even if it is not a risk anymore for anyone. According to the role approach, on the other hand, the risk cannot stop being a risk: if a situation is not risky anymore, the corresponding risk role disappears\footnote{Thus, one might say that an instance of \textit{Risk$^{r}$} is necessarily so, whereas an instance of \textit{Risk$^{d}$} is accidentally so.}. For example, if Paul’s desires change and he doesn't care anymore about being hurt, then the role hazard$_{0}$ disappears: it cannot exist without being a risk.

\subsection{The realizations of the risk-related realizable entities
}

Note that the role approach does not deny the existence of dispositions such as instability$_{0}$: it just denies that this disposition is a risk and construes instead the risk as a role. Thus, the debate between both approaches can be seen as a question whether to map the natural language term “risk” with \textit{Risk$^{d}$} or \textit{Risk$^{r}$}. 

In this approach, the material object ceiling$_{0}$ bears two different realizable entities, the disposition instability$_{0}$ and the role hazard$_{0}$. Their realizations are related but different. If realized, instability$_{0}$ would bring about a BFO:\textit{Process} of collapse in which the materials composing ceiling$_{0}$ loose cohesion and fall down under the action of gravity. In our scenario, this physical process is collapsing$_{0}$, which would happen independently of whether Paul is at risk or not. On the other hand, considering that the integrity of Paul, a normal human being, can be negatively impacted by the falling debris, and that he does not want such an outcome to happen, we identify the realization of hazard$_{0}$ (the risk) as a process of causing damage to Paul. This is a BFO:\textit{Process} named here damaging$_{0}$ in which Paul’s health is negatively affected. Note that collapsing$_{0}$ and damaging$_{0}$ happen approximately at the same time. If we just focus, to make things simple, on the ceiling debris that hit Paul and not those ten meters from him, the falling, physical impact, and biological damage all happen in the same approximate place and time, and with overlapping participants. This does not imply that those processes are identical though. 

Fundamentally, collapsing$_{0}$ and damaging$_{0}$ can be analyzed as sums of spatial changes (change in spatial position of a material entity) and SDC changes respectively, where a SDC change is a  BFO:\textit{Process} that consist in a change of one or several SDCs (following Guarino, Baratella and Guizzardi \cite{b20}, adapted to BFO by Toyoshima and Barton \cite{b21}). One can define collapsing$_{0}$ as the sum of processes of SDC changes (e.g. the structural cohesion of the ceiling that decreases) and processes of spatial changes (the spatial location of the ceiling debris which move down). damaging$_{0}$, on the other hand, might be defined as a change of SDCs characterizing Paul’s health (without taking any specific position on the ontology of health here), which declines while he is hit by those bricks falling down\footnote{Thus, collapsing$_{0}$ and damaging$_{0}$ could be related to process profiles, which formed a subclass of BFO:\textit{Process} until BFO2.0.}. So, those two groups of quality changes co-occur in spatio-temporal regions that are close from each other but different.
Therefore, the role approach admits an additional realizable entity compared to the dispositional approach, namely hazard$_{0}$; and hazard$_{0}$ and instability$_{0}$ are realized in different processes. This is another point of difference with the dispositional approach, which involved only one process, collapsing$_{0}$, and its classification in a specific subclass.

\subsection{Merely possible processes}
A central tenet of all discourses about risk is possibility. We have identified three processes so far: Paul actively entertaining a desiring process of not being hurt, collapsing$_{0}$, and damaging$_{0}$. However, even if none of these three processes would happen - in particular, if instability$_{0}$ and hazard$_{0}$ would have no realization, the whole situation is a risky one. How can we then express this unrealized possibility in the ontology?

A first solution, following \cite{b22} would be to state that, if realized, instability$_{0}$ is realized in an instance of the class \textit{ceiling$_{0}\_$collapsing} (which is, here too, a defined class and not a universal). This can be formalized with the axiom (O1): \\

(O1) instability$_{0}$ realized$\_$in only \textit{ceiling$_{0}\_$collapsing} \\

However, a challenge with (O1) is that according to BFO, only classes that have an instance existing in our world (be it in the past, present, and/or future) can be introduced.
 
This means that the class \textit{ceiling$_{0}\_$collapsing} might not exist, as there might be no collapse of ceiling$_{0}$ in the history of our world, and thus should not be introduced.

A possible way to deal with this problem is to formalize an axiom (O2) slightly differently by introducing instead the universal \textit{Collapsing}\footnote{The authors thank Barry Smith for suggesting this idea in a private discussion. Note that ‘has$\_$participant value ceiling$_{0}$’ refers, in OWL Manchester syntax, to the class of entities that have as participant ceiling$_{0}$}.\\

(O2) instability$_{0}$ realized$\_$in only (\textit{Collapsing} and has$\_$participant value ceiling$_{0}$) \\

For sure, some material entities have collapsed in the history of our world; therefore, the universal \textit{Collapsing} exists according to BFO. Thus, the axiom O2 and the entities mentioned there can fit into BFO’s metaphysical framework. This strategy can be straightforwardly adapted to the processes of Paul being damaged or Paul actively desiring not to be hurt.

   \section{Conative theories or risk}
   
\subsection{A knowledge agnostic approach}

According to both the dispositional and the role approach, risks come into existence because some unwanted outcomes that would negatively impact an agent might occur. So far, we have modeled this condition with a process of desiring entertained by an agent, in our case Paul, or some other conative mental state. Paul might walk under the ceiling with or without knowledge of its instability, holding a desire of self-preservation in both cases. Many publications on risk analysis focus on what an agent knows or forecasts about possible situations and which actions he takes or avoids given their beliefs \cite{b23}. Some other studies consider whether some of these actions are rational or ethical \cite{b24}. Here, we only focus on the metaphysics of risk, and we argue that the knowledge of Paul is not relevant for determining what is a risk for him: when ceiling$_{0}$ is unstable and Paul stands below, the risk is there, independently of his knowledge of the situation. Similarly, the appropriate ethical response to a risk has no bearing on the metaphysics of risk. Therefore, epistemological or ethical features are largely irrelevant to our current investigation. Invoking desires allows us to have a belief-agnostic approach in modeling risk, which means in our scenario that the beliefs and knowledge of Paul are irrelevant to determine what is risky. Paul’s beliefs and course of actions could determine whether he is a good risk assessor and his degree of risk aversion, but this is a matter for decision theory or psychology of decision making. Additionally, as already hinted in the section on merely possible processes, Paul could be ignorant about his own desires, by having a dispositional desire that remains unrealized in explicit mental processes. In this we follow a tradition analyzing many mental entities as dispositions, including beliefs, desires, and intentions \cite{b25}. The desire disposition can remain latent for some time and can be realized in mental desiring processes, typically conscious, but that might be unconscious \cite{b26}. If Paul is sleeping or comatose under ceiling$_{0}$, there is still a risk that it would collapse on him, even without a conscious process of desiring safety, and independently of his (dispositional) beliefs.

\subsection{Desires and aversions}
We consider here desires as mental dispositions directed towards some objects. Moreover, given that risks concern scenarios in which some damage or problem might occur, it seems the desires relevant for risk scenarios are connected to something that might not happen. What seems easy from a linguistic point of view is more difficult from an ontological point of view. It is classically problematic for ontology to deal with negation or inexistence \cite{b27}.

First of all, it is relevant to distinguish between desiring not-x and not desiring x. The first case concerns an active process of the mind wishing that a certain state of affairs would not obtain. In the second case, a mental process is lacking: even stones can be said to not desire something, whereas they certainly cannot desire not-x.

When risk roles are realized in processes, those processes are contrary to an agent’s desires. But realist ontologies do not accept things such as negative processes or negative objects - or more generally negative entities (see also \cite{b28} and \cite{b29}). The negation is rather embedded in a peculiar mental entity opposite to desiring. As in the mental realm love is the opposite of hate and hope is the opposite of despair, we can consider the mental attitude of aversion, which is the opposite of desiring, and might satisfy statements such as: “if an agent desires x, then he does not avert x”. Similarly to desires, aversions are dispositions that can be realized in mental processes - in this case, processes of averting. This therefore dispenses us from introducing problematic entities as negations in our model, and only connects existing entities together. To be more precise, mental attitudes like aversions and desires are connected to what they represent by what is usually called a relation of “intentionality” or “aboutness” \cite{b30}, which we will call here “is$\_$directed$\_$towards”.

\subsection{Intentionality towards Risk Realizations}

Having tackled the negativity of certain mental conative entities like aversion doesn’t settle the issue of which real-world entity does Paul avert - that is, towards which entity is the mental attitude of aversion directed. Entities that Paul might avert include ceiling$_{0}$, its disposition instability$_{0}$ the process collapsing$_{0}$ or the process damaging$_{0}$. What Paul fundamentally averts is to be bodily hurt, so he would avert the process damaging$_{0}$, which would realize the risk hazard$_{0}$. However, his mental attitude of aversion would exist even if hazard$_{0}$ was not realized in the process damaging$_{0}$. Exploiting a similar solution to (O2) that we have seen with merely possible processes, we can say that aversion$_{0}$ can only have an intentionality relation with something that is a damaging process, if any, and that this process has Paul as its passive participant, since he would take the damage\footnote{The authors thank John Beverley for this suggestion about intentional objects in OWL following BFO.}. In OWL: \\

(O3) aversion$_{0}$ is$\_$directed$\_$towards only (\textit{Damaging} and
has$\_$participant value Paul) \\

In this way, the sentence (O3) can be added in an ontology and be true even if there were no specific instances of \textit{Damaging} involving Paul at the time the mental attitude of aversion was held by Paul.

Given this mental aversion for a generic damaging process, we can actually model another situation, in which ceiling$_{0}$ has been heavily painted with a poisonous paint which could impact Paul’s health if he passed close to it, given the paint can release volatile particles in the air. The same object, ceiling$_{0}$, and the same latent disposition of aversion, can explain several risks in this way. In the scenario in which the ceiling is both unstable and poisonous, aversion$_{0}$ can be intentionally directed to both harmful realizations of damaging by collapse and damaging by poisoning, since averting bodily damage is generic enough to comprise them both. For sure, we could model another agent, more selective than Paul, who only averts being damaged by the collapse of something from above, a very specific process, and so the poisonous ceiling would not lead to a risk for this selective agent, since they would not care. The lesson to be learned is that the intentional direction of aversion determines what is a risk or not.

Instead of averting the realization of a risk role, consider if we had said that Paul averted a process of collapsing of ceiling$_{0}$ which is the realization of a disposition, not of a risk role: such a process could happen even when Paul was miles away from ceiling$_{0}$, with no possibility of being hurt by it. He might actually feel some schadenfreude and rejoice at the sight of a collapsing building, for all we know. Maybe Paul likes watching videos of collapsing buildings, so this kind of process is not averted in itself, but only because it brings about a process of damaging Paul’s health. This is an important distinction between the realizations of instances of \textit{Risk$^{d}$} and \textit{Risk$^{r}$}, as well as a virtue of modeling risk as the latter, as it grasps what is really at the core of the conative states of an agent.

Risk roles are externally grounded properties in the sense that they exist in virtue of something else than the entity they inhere in. The risk hazard$_{0}$ is not there just because of ceiling$_{0}$, but because of some complex situation of an agent desiring something, which ontologically grounds it. With this in mind and having found all the suitable entities to express a situation of risk, we can finally clarify the “context” that we introduced earlier and understand better the entities and the connections present there. The objective side of risk that we have modeled first, with physical interactions between the parties involved, has now been articulated with the subjective side foreshadowed at the beginning. Paul’s mental disposition aversion$_{0}$, which is intentionally directed towards a class of damaging processes, grounds the existence of the risk role hazard$_{0}$ in ceiling$_{0}$.

Which intentional objects are targeted by the mental attitude determines which and how many risks there are, so outside of our example, if Paul also cared for its clothes and not just his health, which could be ruined by the falling debris of ceiling$_{0}$, a second risk, let’s say hazard$_{1}$, would pop out into existence. Such risk would be grounded in a new mental attitude such as aversion$_{1}$, that would be intentionally directed towards a class of processes of clothes being ruined.

\section{Towards a definition of risk}

So far, we have illustrated our framework with some specific instances in a very particular situation that enabled us to illustrate the agent-relativity of risk. We can now present the framework of risk as a role in Fig. 2. Note that we depicted the case in which ceiling$_{0}$ actually falls during the collapsing$_{0}$ process and hazard$_{0}$ is realized in damaging$_{0}$, but such processes might never occur. Moreover, to improve readibility, the relations participates$\_$in and is$\_$directed$\_$towards have been omitted.

\begin{figure}[htbp]
\begin{center}
\includegraphics[scale=0.160]{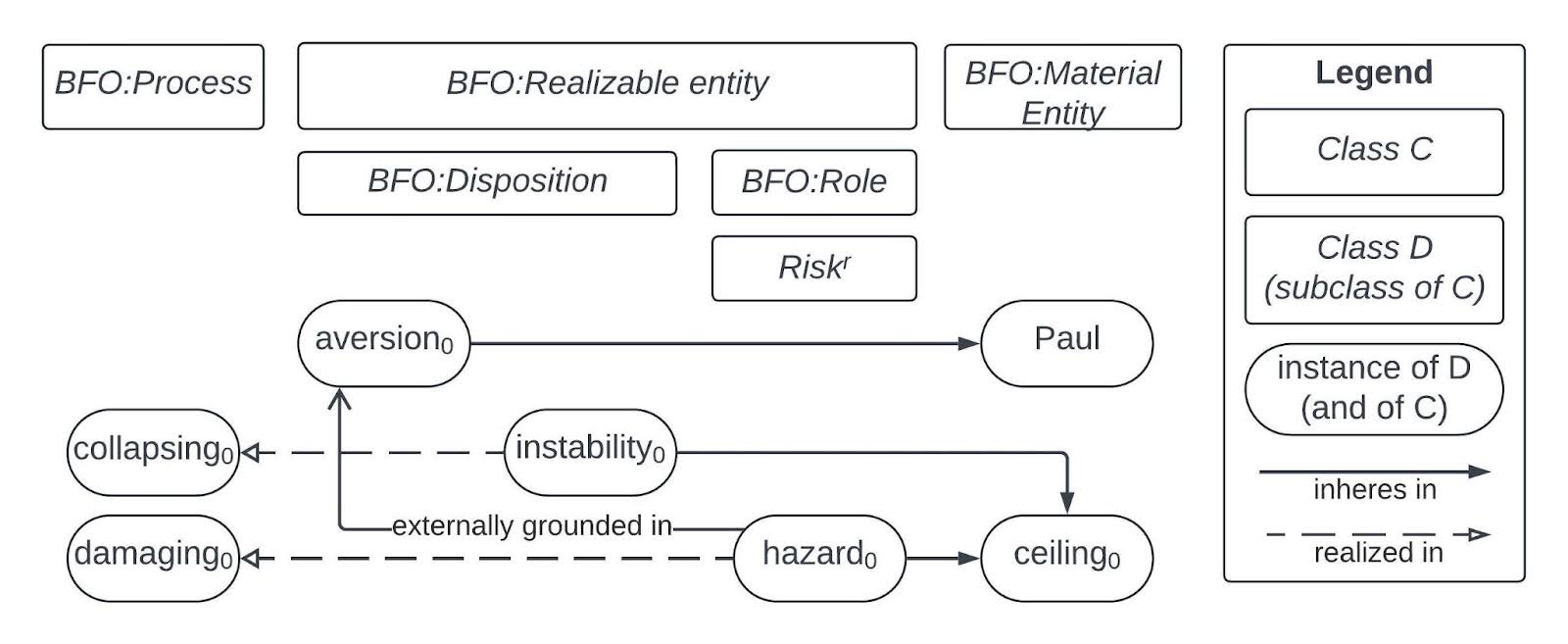}
\end{center}
\caption{\textit{Risk$^{r}$} as a BFO:\textit{Role}}
\label{fig}
\end{figure}

Now, we abstract away from this particular scenario and its instances to address risk more generally. The bearer of a risk might be a BFO:\textit{Object} (think about a knife), or other kinds of BFO:\textit{Material entity}, such as a body of water (e.g. in a flood) or instances of more specific BFO categories (consider the risk that a flock of birds, which is a BFO:\textit{Object} aggregate, would attack a human). It is also important to consider that bearers of risk must be causally effective in producing the unwanted consequence an agent would avert, which supports the choice of a BFO:\textit{Material entity} as the bearer of a risk, and that such bearer must also have some specific BFO:\textit{Disposition}. Finally, the entity that averts the realization of the risk could be an agent, like in our example, or a group of agents, like in the case of an entire population threatened by war (we do not discuss here whether such a group of agents would form an agent at another level).

Now, are these entities and relations enough to define risk? Arguably, some elements in our definition of risk might be too specific and unnecessary. Indeed, some risks might relate to teleological aspects that are more general than agents aversion - consider e.g. the risk for a plant to die when set aflame: burning would be the physical process involved, which is bad only within the perspective of the goal of the plant to stay alive. But it is debatable whether plants can desire or avert anything. Therefore, the conative approach might have to be generalized to also comprise this plant example.

Another element hinting towards a more general definition of risk, that we cannot elaborate here, is the possibility for non-material entities to bear risks. It is still to be determined if plausible risky situations like a financial loss or a cyber-attack only involve material entities, or if the source of damage could have a different nature.

Given these considerations, we propose the following sufficient (but not necessary) conditions for risk: \\

\noindent \textit{If}: 
(i) r is a role \& \\
(ii) r is externally grounded in an aversion that an agent or group of agents have towards the possible realizations of the role, \\
\textit{then} r is a risk. \\
   
In case risk will be discovered to be only agentive and the risk bearer always material, then the condition above would actually be necessary and sufficient  and constitute a definition. This shall be investigated in future work.

\section{Conclusion and future work}

To conclude, we have explored a novel way of understanding risk as a BFO:\textit{Role}, after considering a similar approach which categorized it as a BFO:\textit{Disposition}. In order to complete our modeling, we made use of realizable entities in BFO, opening the way for a future ontology of risks as roles. We proposed a sufficient condition for risks: future work should inquire whether this is also a necessary condition, or how it could be refined into a necessary and sufficient condition. A limitation of this exploratory investigation was its general character, and the lack of competency questions from domain experts to evaluate the model. Future work should also clarify the nature of a damaging process, what entities can be damaged and what is the output of damaging, since it might involve more considerations on negations, absences and realizable entities. Our risk ontology will need to be connected with risk probabilities (as started in \cite{b9} and \cite{b31}) and other relevant quantities: we merely covered qualitative possibilities so far, but the quantification of risks often involves probability measures and severity degrees. The ontology of risk propagation should also be investigated, where material entities can acquire and lose risk roles in a chain of causal processes. Furthermore, it will be important to evaluate this ontology of risk with real data and show how it can structure the information present in risk registers for better clarity and interoperability.



\begin{thebibliography}{00}

\bibitem{b1} M. C. Leva, N. Balfe, B. McAleer, and M. Rocke, “Risk registers: Structuring data collection to develop risk intelligence,” Safety Science, vol. 100, pp. 143–156, Dec. 2017, doi: 10.1016/j.ssci.2017.05.009.

\bibitem{b2} M. Andretta, “Some Considerations on the Definition of Risk Based on Concepts of Systems Theory and Probability,” Risk Analysis, vol. 34, no. 7, pp. 1184–1195, Jul. 2014, doi: 10.1111/risa.12092.

\bibitem{b3} P. J. Blokland and G. L. Reniers, “The Concepts of Risk, Safety, and Security: A Fundamental Exploration and Understanding of Similarities and Differences,” pp. 9–16, 2020, doi: 10.1007/978-3-030-47229-0\_2.

\bibitem{b4} M. Leitch, “ISO 31000:2009—The New International Standard on Risk Management,” Risk Analysis, vol. 30, no. 6, pp. 887–892, Jun. 2010, doi: 10.1111/j.1539-6924.2010.01397.x.

\bibitem{b5} N. Guarino, “Understanding, building and using ontologies,” International Journal of Human-Computer Studies, vol. 46, no. 2–3, pp. 293–310, Feb. 1997, doi: 10.1006/ijhc.1996.0091.

\bibitem{b6} R. Arp, B. Smith, and A. D. Spear, Building ontologies with Basic Formal Ontology. Cambridge, Massachusetts: The MIT Press, 2015.

\bibitem{b7} A. Iliadis, “The Tower of Babel problem: making data make sense with Basic Formal Ontology,” Online Information Review, vol. 43, no. 6, pp. 1021–1045, Jan. 2019, doi: 10.1108/OIR-07-2018-0210.

\bibitem{b8} T. P. Sales, F. Baião, G. Guizzardi, J. P. A. Almeida, N. Guarino, and J. Mylopoulos, “The Common Ontology of Value and Risk,” in Conceptual Modeling: 37th International Conference, ER 2018, Xi’an, China, October 22–25, 2018, J. C. Trujillo, K. C. Davis, X. Du, Z. Li, T. W. Ling, G. Li, and M. L. Lee, Eds., Cham: Springer International Publishing, 2018, pp. 121–135. doi: 10.1007/978-3-030-00847-5\_11.

\bibitem{b9} A. Barton, L. Jansen, A. Rosier, and J.-F. Ethier, “What is a risk? A formal representation of risk of stroke for people with atrial fibrillation,” in Proceedings of the 8th International Conference on Biomedical Ontology (ICBO 2017), Newcastle, UK: CEUR Workshop Proceedings, 2018, pp. 1–6. [Online]. Available: http://ceur-ws.org/Vol-2137/

\bibitem{b10} Kim, Jin Hyun and Park, Dal Jae, “A Study on the Review of Risk Concepts,” in Journal of the Korean Society of Safety, Oct. 2013, pp. 90–96. doi: 10.14346/JKOSOS.2013.28.6.090.

\bibitem{b11} T. Aven et al., “Society for Risk Analysis Glossary,” Society for Risk Analysis, Aug. 2018. [Online]. Available: https://backend.orbit.dtu.dk/ws/portalfiles/portal/377037938/Society


\bibitem{b12} S. O. Hansson, “Risk,” in The Stanford Encyclopedia of Philosophy, Spring 2014., E. N. Zalta, Ed., Metaphysics Research Lab, Stanford University, 2014. [Online]. Available: https://plato.stanford.edu/archives/spr2014/entries/risk/


\bibitem{b13} B. Smith, “Introduction to the Logic of Definitions,” in International Workshop on Definitions in Ontologies, organized in conjunction with the Fourth International Conference on Biomedical Ontology (ICBO), Montreal, July 7, 2013, (CEUR, 1061), 2013, pp. 1–2.


\bibitem{b14} A. Varzi, “Mereology,” The Stanford Encyclopedia of Philosophy. Metaphysics Research Lab, Stanford University, 2019. Accessed: Nov. 17, 2023. [Online]. Available: https://plato.stanford.edu/archives/spr2019/entries/mereology/

\bibitem{b15} M. Rabenberg et al., “Grounding Realizable Entities,” Apr. 30, 2024, arXiv: arXiv:2405.00197. doi: 10.48550/arXiv.2405.00197.

\bibitem{b16} R. Arp and B. Smith, “Function, Role, and Disposition in Basic Formal Ontology,” Nat Prec, Jun. 2008, doi: 10.1038/npre.2008.1941.1.

\bibitem{b17} O. Grenier and A. Barton, “Une Ontologie Dispositionnelle du Risque,” Lato Sensu: Revue de la Société de Philosophie des Sciences, vol. 8, no. 2, pp. 58–69, 2021, doi: 10.20416/lsrsps.v8i2.6.


\bibitem{b18} B. Smith and W. Ceusters, “Ontological realism: A methodology for coordinated evolution of scientific ontologies,” Applied Ontology, vol. 5, no. 3–4, pp. 139–188, Jan. 2010, doi: 10.3233/AO-2010-0079.

\bibitem{b19} F. Toyoshima, A. Barton, L. Jansen, and J.-F. Ethier, “Towards a Unified Dispositional Framework for Realizable Entities,” in Formal Ontology and Information Systems (FOIS), Bozen-Bolzano, Italy, Sep. 2021.


\bibitem{b20} N. Guarino, R. Baratella, and G. Guizzardi, “Events, their names, and their synchronic structure,” Applied Ontology, vol. 17, no. 2, pp. 249–283, May 2022, doi: 10.3233/AO-220261.

\bibitem{b21} F. Toyoshima and A. Barton, “Two Approaches to the Identity of Processes in BFO,” in Formal Ontology in Information Systems. Proceedings of the 11th International Conference (FOIS 2020), B. Broadaric and F. Neuhaus, Eds., Bolzano, Italy: IOS Press, 2020, pp. 140–154. [Online]. Available: https://hal.science/hal-04301542
  

\bibitem{b22} J. Röhl and L. Jansen, “Representing Dispositions,” Journal of Biomedical Semantics, vol. 2, no. 4, 2011. doi: 10.1186/2041-1480-2-S4-S4.

\bibitem{b23} S. Roeser, Handbook of Risk Theory: Epistemology, Decision Theory, Ethics, and Social Implications of Risk. Springer Science \& Business Media, 2012.

\bibitem{b24} S. Hansson, The Ethics of Risk: Ethical Analysis in an Uncertain World. Springer, 2013.

\bibitem{b25} F. Toyoshima, A. Barton, and O. Grenier, “Foundations for an ontology of belief, desire and intention,” in Formal Ontology in Information Systems. Proceedings of the 11th International Conference (FOIS 2020), B. Broadaric and F. Neuhaus, Eds., Bolzano, Italy: IOS Press, 2020, pp. 140–154. doi: 10.3233/FAIA200667.

\bibitem{b26} J. Hastings, B. Smith, W. Ceusters, M. Jensen, and K. Mulligan, “Representing mental functioning: Ontologies for mental health and disease,” in ICBO 2012: 3rd International Conference on Biomedical Ontology, University of Graz, 2012, pp. 1–5.

\bibitem{b27} W. Ceusters, P. Elkin, and B. Smith, “Negative findings in electronic health records and biomedical ontologies: A realist approach,” International Journal of Medical Informatics, vol. 76, pp. S326–S333, Dec. 2007, doi: 10.1016/j.ijmedinf.2007.02.003.

\bibitem{b28} R. Demos, “A Discussion of a Certain Type of Negative Proposition,” Mind, vol. 26, no. 102, pp. 188–196, 1917.

\bibitem{b29} S. Mumford, “Negation and Denial,” in Cambridge Handbook of the Philosophy of Language, P. Stalmaszczyk, Ed., Cambridge University Press, 2021.

\bibitem{b30} P. Jacob, “Intentionality,” The Stanford Encyclopedia of Philosophy. Metaphysics Research Lab, Stanford University, 2023. [Online]. Available: https://plato.stanford.edu/archives/spr2023/entries/intentionality/

\bibitem{b31} A. Barton, A. Burgun, and R. Duvauferrier, “Probability assignments to dispositions in ontologies,” in Formal Ontology and Information Systems. Proceedings of the Seventh International Conference (FOIS 2012), M. Donnelly and G. Guizzardi, Eds., Amsterdam: IOS Press, 2012, pp. 3–14., 2012, pp. 3–14.  

\end{thebibliography}
\end{document}